\documentclass[a4paper, 10pt, conference]{ieeeconf}
\usepackage[utf8]{inputenc}
\IEEEoverridecommandlockouts
\usepackage{amsmath} 
\usepackage{amssymb}  
\usepackage{algorithm}
\usepackage[noend]{algpseudocode}

\usepackage{bm}
\DeclareMathOperator*{\argmax}{arg\,max}
 
\newcommand{\minus}{\scalebox{0.75}[1.0]{$-$}}

\usepackage{booktabs}
\usepackage[pdftex]{graphicx}
\usepackage{subcaption}
\usepackage{hyperref}
\usepackage[stable]{footmisc}
\usepackage{balance}
\usepackage{multirow}

\usepackage[usenames,dvipsnames]{color}

\definecolor{wheat}{rgb}{0.96,0.87,0.70}
\definecolor{tommy}{rgb}{0.8,0.8,1}
\definecolor{cj}{rgb}{1,0.8,0.8}

\title{\LARGE \bf
Reinforcement Learning with Uncertainty Estimation\\for Tactical Decision-Making in Intersections
}

\author{Carl-Johan Hoel\textsuperscript{*,1,3,4}, Tommy Tram\textsuperscript{*,2,3,4} and Jonas Sj\"oberg\textsuperscript{3} 
\thanks{\textsuperscript{*} Both authors contributed equally to this work.}
\thanks{This work was partially supported by the Wallenberg Artificial Intelligence, Autonomous Systems, and Software Program (WASP), funded by the Knut and Alice Wallenberg Foundation, and partially by Vinnova FFI.}
\thanks{\textsuperscript{1} Affiliated with Volvo Group, Gothenburg, Sweden.}%
\thanks{\textsuperscript{2} Affiliated with Zenuity AB, Gothenburg, Sweden.}%
\thanks{\textsuperscript{3} Affiliated with Chalmers University of Technology, Gothenburg, Sweden.
        {\tt \{carl-johan.hoel,tommy.tram,jonas.sjoberg\}
        @chalmers.se}}%
\thanks{\textsuperscript{4} Affiliated with AI Innovation of Sweden, Gothenburg, Sweden.}%
}

\begin{document}

\maketitle
\thispagestyle{empty}
\pagestyle{empty}

\begin{abstract}


This paper investigates how a Bayesian reinforcement learning method can be used to create a tactical decision-making agent for autonomous driving in an intersection scenario, where the agent can estimate the confidence of its recommended actions. An ensemble of neural networks, with additional randomized prior functions (RPF), are trained by using a bootstrapped experience replay memory. The coefficient of variation in the estimated $Q$-values of the ensemble members is used to approximate the uncertainty, and a criterion that determines if the agent is sufficiently confident to make a particular decision is introduced. The performance of the ensemble RPF method is evaluated in an intersection scenario, and compared to a standard Deep Q-Network method. It is shown that the trained ensemble RPF agent can detect cases with high uncertainty, both in situations that are far from the training distribution, and in situations that seldom occur within the training distribution. In this study, the uncertainty information is used to choose safe actions in unknown situations, which removes all collisions from within the training distribution, and most collisions outside of the distribution.

\end{abstract}

\section{Introduction}







To make safe, efficient, and comfortable decisions in intersections is one of the challenges of autonomous driving. 
A decision-making agent needs to handle a diverse set of intersection types and layouts, interact with other traffic participants, and consider uncertainty in sensor information. 
The fact that around 40\% of all traffic accidents during manual driving occur in intersections indicates that decision-making in intersections is a complex task~\cite{NHTSA}.
To manually predict all situations that can occur and tailor a suitable behavior is not feasible. Therefore, a data-driven approach that can learn to make decisions from experience is a compelling approach. A desired property of such a machine learning approach is that it should also be able to indicate how confident the resulting agent is about a particular decision.

Reinforcement learning (RL) provides a general approach to solve decision-making problems~\cite{Sutton2018}, and could potentially scale to all types of driving situations. Promising results have been achieved in simulation by applying a Deep Q-Network (DQN) agent to intersection scenarios~\cite{Isele2018,Tram2018}, and highway driving~\cite{Wang2018,Hoel2018}, or a policy gradient method to a lane merging situation~\cite{Shalev2016}. Some studies have trained an RL agent in a simulated environment and then deployed the agent in a real vehicle~\cite{Pan2017, Bansal2018}, and for a limited case, trained the agent directly in a real vehicle~\cite{Kendall2017}.

Generally, a fundamental problem with the RL methods in previous work is that the trained agents do not provide any confidence measure of their decisions.
For example, if an agent that was trained for a highway driving scenario would be exposed to an intersection situation, it would still output a decision, although it would likely not be a good one. A less extreme example involves an agent that has been trained in an intersection scenario with nominal traffic, and then faces a speeding driver. 
McAllister et al. further discuss the importance of estimating the uncertainty of decisions in autonomous driving~\cite{McAllister2017}.

A common way of estimating uncertainty is through Bayesian probability theory~\cite{Kochenderfer2015}. Bayesian deep learning has previously been used to estimate uncertainty in autonomous driving for image segmentation~\cite{Kendall2017} and end-to-end learning~\cite{Michelmore2018}. Dearden et al. introduced Bayesian approaches to RL that balances the trade off between exploration and exploitation~\cite{Dearden1998}. In recent work, this approach has been extended to deep RL, by using an ensemble of neural networks~\cite{Osband2018}. However, these studies focus on creating an efficient exploration method for RL, and do not provide a confidence measure for the agents' decisions.

This paper investigates an RL method that can estimate the uncertainty of the resulting agent's decisions, applied to decision-making in an intersection scenario. The RL method uses an ensemble of neural networks with randomized prior functions that are trained on a bootstrapped experience replay memory, which gives a distribution of estimated $Q$-values (Sect.~\ref{sec:approach}). The distribution of $Q$-values is then used to estimate the uncertainty of the recommended action, and a criterion that determines the confidence level of the agent's decision is introduced (Sect.~\ref{sec:safetyCriterion}). The method is used to train a decision-making agent in different intersection scenarios (Sect.~\ref{sec:implementation}), in which the results show that the introduced method outperforms a DQN agent within the training distribution. The results also show that the ensemble method can detect situations that were not present in the training process, and thereby choose safe fallback actions in such situations (Sect.~\ref{sec:results}).
Further characteristics of the introduced method is discussed in Sect.~\ref{sec:discussion}.
This work is an extension to a recent paper, where we introduced the mentioned method, but applied to a highway driving scenario~\cite{Hoel2020}.

\section{Approach}
\label{sec:approach}
This section gives a brief introduction to RL, describes how the uncertainty of an action can be estimated by an ensemble method, and introduces a measure of confidence for different actions. Further details on how this approach was applied to driving in an intersection scenario follows in Sect.~\ref{sec:implementation}.

\subsection{Reinforcement learning}

Reinforcement learning is a branch of machine learning, where an agents explores an environment and tries to learn a policy $\pi(s)$ that maximizes the future expected return, based on the agent's experiences~\cite{Sutton2018}. The policy determines which action $a$ to take in a given state $s$. The state of the environment will then transitions to a new state $s'$ and the agent receives a reward $r$. A Markov Decision Process (MDP) is often used to model the reinforcement learning problem. An MDP is defined by the tuple $( \mathcal{S}, \mathcal{A}, T, R, \gamma)$, where $\mathcal{S}$ is the state space, $\mathcal{A}$ is the action space, $T$ is a state transition model, $R$ is a reward model, and $\gamma$ is a discount factor. At each time step $t$, the agent tries to maximize the future discounted return
\begin{align}
    R_t = \sum_{k=0}^\infty \gamma^k r_{t+k}.
\end{align}

In a value-based branch of RL called $Q$-learning~\cite{Watkins1992}, the objective of the agent is to learn the optimal state-action value function $Q^*(s,a)$. This function is defined as the expected return when the agent takes action $a$ from state $s$ and then follow the optimal policy $\pi^*$, i.e.,
\begin{align}
    Q^*(s,a) = \max_\pi \mathbb{E} \left[R_t | s_t = s, a_t = a, \pi\right].
\end{align}
The $Q$-function can be estimated by a neural network with weights $\theta$, i.e., $Q(s,a) \approx Q(s,a;\theta)$. The weights are optimized by minimizing the loss function
\begin{align}
    L(\theta) = \mathbb{E}_M \Big[ (r + \gamma \max_{a'} Q(s',a';\theta^-)
    - Q(s,a;\theta) )^2 \Big],
    \label{eq:loss}
\end{align}
which is derived from the Bellman equation. The loss is obtained from a mini-batch $M$ of training samples, and $\theta^-$ represents the weights of a target network that is updated regularly. More details on the DQN algorithm are presented by Mnih et al.~\cite{Mnih2015}.

\subsection{Bayesian reinforcement learning}

One limitation of the DQN algorithm is that only the maximum likelihood estimate of the $Q$-values is returned. The risk of taking a particular action can be approximated as the variance in the estimated $Q$-value~\cite{Garcia2015}. One approach to obtain a variance estimation is through statistical bootstrapping~\cite{Efron1982}, which has been applied to the DQN algorithm~\cite{Osband2016}. The basic idea is to train an ensemble of neural network on different subsets of the available replay memory. The ensemble will then provide a distribution of $Q$-values, which can be used to estimate the variance. Osband et al. extended the ensemble method by adding a randomized prior function (RPF) to each ensemble member, which gives a better Bayesian posterior~\cite{Osband2018}. The $Q$-values of each ensemble member $k$ is then calculated as the sum of two neural networks, $f$ and $p$, with equal architecture, i.e.,
\begin{align}
    Q_k(s,a) = f(s,a;\theta_k) + \beta p(s,a;\hat{\theta}_k).
\end{align}
Here, the weights $\theta_k$ of network $f$ are trainable, and the weights $\hat{\theta}_k$ of the prior network $p$ are fixed to the randomly initialized values. A parameter $\beta$ scales the importance of the networks. With the two networks, the loss function in Eq.~\ref{eq:loss} becomes
\begin{align}
    \label{eq:loss_boot}
    L(\theta_k) = \mathbb{E}_M \Big[ & (r + \gamma \max_{a'} (f_{\theta^-_k}+\beta p_{\hat{\theta}_k})(s',a') \nonumber \\
    & - (f_{\theta_k}+ \beta p_{\hat{\theta}_k})(s,a) )^2 \Big].
\end{align} 

Algorithm~\ref{alg:training} outlines the complete ensemble RPF method, which was used in this study. An ensemble of $K$ trainable and prior neural networks are first initialized randomly. Each ensemble member is also assigned a separate experience replay memory buffer $m_k$ (although in a practical implementation, the replay memory can be designed in such a way that it uses negligible more memory than a shared buffer). For each new training episode, a uniformly sampled ensemble member, $\nu \sim \mathcal{U}\{1,K\}$, is used to greedily select the action with the highest $Q$-value. This procedure handles the exploration vs. exploitation trade-off and corresponds to a form of approximate Thompson sampling. Each new experience $e = (s_i, a_i, r_i, s_{i+1})$ is then added to the separate replay buffers $m_k$ with probability $p_\mathrm{add}$. Finally, the trainable weights of each ensemble member are updated by uniformly sample a mini-batch $M$ of experiences and using stochastic gradient descent (SGD) to backpropagate the loss of Eq.~\ref{eq:loss_boot}.

\begin{algorithm}[!t]
    \caption{Ensemble RPF training process}\label{alg:training}
    \begin{algorithmic}[1]
        \For{$k \gets 1$ to $K$}
            \State Initialize $\theta_k$ and $\hat{\theta}_k$ randomly
            \State $m_k \gets \{\}$
        \EndFor
        \State $i \gets 0$
        \While{networks not converged}
            \State $s_i \gets $ initial random state
            \State $\nu \sim \mathcal{U}\{1,K\}$
            \While{episode not finished}
                \State $a_i \gets \argmax_{a} Q_\nu(s_i,a)$
                \State $s_{i+1}, r_i \gets $ \Call{StepEnvironment}{$s_i, a_i$}
                \For{$k \gets 1$ to $K$}
                   \If{$p \sim \mathcal{U}(0,1) < p_\mathrm{add}$}
                      \State $m_k \gets m_k \cup \{(s_i, a_i, r_i, s_{i+1})\}$
                   \EndIf
                   \State $M \gets $ sample mini-batch from $m_k$
                   \State update $\theta_k$ with SGD and loss $L(\theta_k)$
                \EndFor
                \State $i \gets i + 1$
            \EndWhile
        \EndWhile
    \end{algorithmic}
\end{algorithm}

\subsection{Confidence criterion}
\label{sec:safetyCriterion}

The agent's uncertainty in choosing different actions can be defined as the coefficient of variation\footnote{Ratio of the standard deviation to the mean.} $c_\mathrm{v}(s,a)$ of the $Q$-values of the ensemble members.  
In a previous study, we introduced a confidence criterion that disqualifies actions with $c_\mathrm{v}(s,a) > c_\mathrm{v}^\mathrm{safe}$, where $c_\mathrm{safe}$ is a hard threshold~\cite{Hoel2020}. 
The value of the threshold should be set so that $(s,a)$ combinations that are contained in the training distribution are accepted, and those which are not will be rejected. This value can be determined by observing values of $c_\mathrm{v}$ in testing episodes within the training distribution, see Sect.~\ref{sec:resultsWithinDistribution} for further details.


When the agent is fully trained (i.e., not during the training phase), the policy chooses actions by maximizing the mean of the $Q$-values of the ensemble members, with the restriction $c_\mathrm{v}(s,a) < c_\mathrm{v}^\mathrm{safe}$, i.e.,
\begin{equation}
    \begin{aligned}
        \argmax_{a} \frac{1}{K} \sum_{k=1}^K Q_k(s,a),\\
        \textrm{s.t.} \quad c_\mathrm{v}(s,a) < c_\mathrm{v}^\mathrm{safe}.
    \end{aligned}
\end{equation}
In a situation where no possible action fulfills the confidence criterion, a fallback action $a_\mathrm{safe}$ is chosen.

\section{Implementation}
\label{sec:implementation}


The ensemble RPF method, which can obtain an uncertainty estimation of different actions, is tested on different intersection scenarios. In this study, the uncertainty information is used to reject unsafe actions and reduce the number of collisions. 
This section describes how the simulation of the scenarios is set up, how the decision-making problem is formulated as an MDP, the architecture of the neural networks, and the details on how the training is performed.

\subsection{Simulation setup}
\label{sec:simulationModel}

\begin{figure}[t]
	\centering
	\begin{subfigure}[t]{0.4\columnwidth}
		\centering
		\includegraphics[width=\columnwidth]{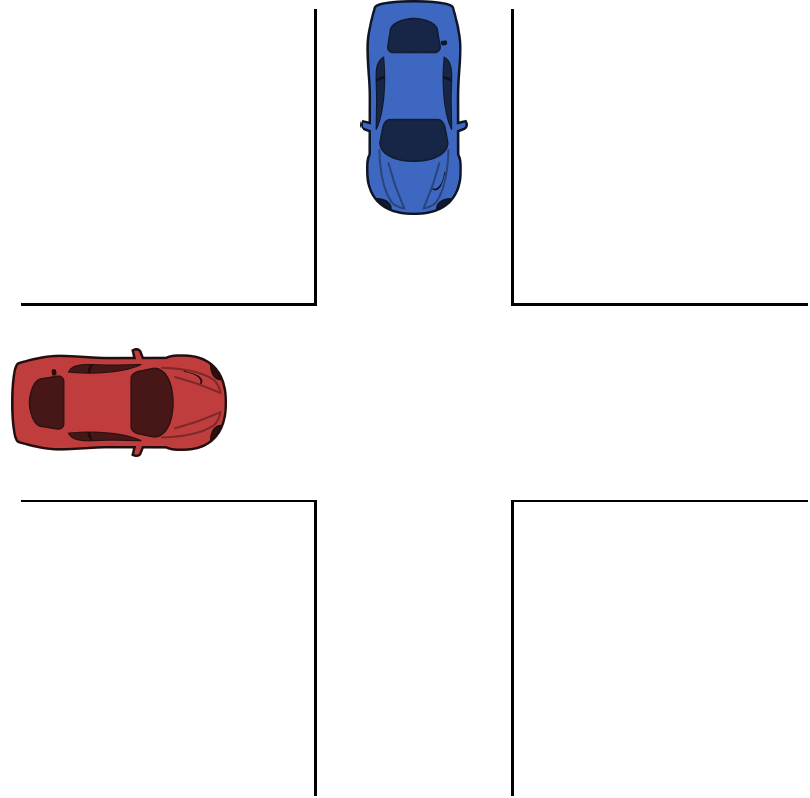}
		
	\end{subfigure}
	\hfill
	\rule{0.5px}{100px}
	\hfill
	\begin{subfigure}[t]{0.45\columnwidth}
		\centering
		\includegraphics[width=\columnwidth]{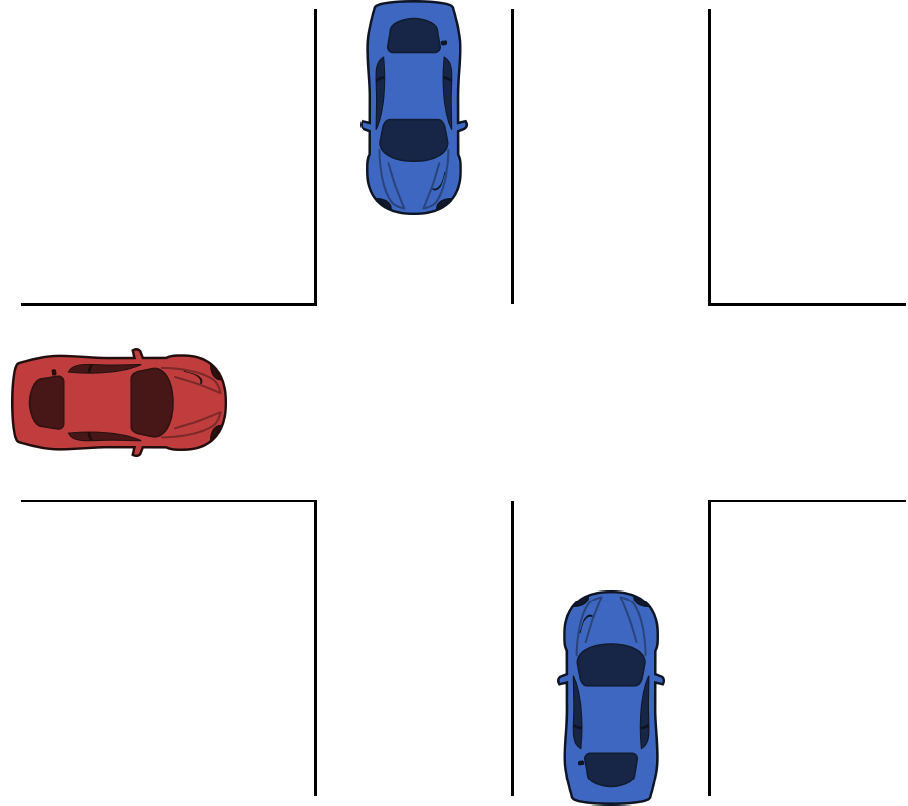}
	\end{subfigure}
	
	\caption{The two intersection scenarios that were used in this study; single directional to the left and bidirectional to the right.}
	\label{fig:scenarios}
\end{figure}

The simulated environment consists of different intersection scenarios, and is based on a previous study~\cite{tram2019}. For completeness, an overview is presented here. Each episode starts by randomly selecting a single or bi-directional intersection, shown in Fig. \ref{fig:scenarios}, and placing the ego vehicle to the left with a random distance $p^{\mathrm{c},j}_\mathrm{e}$ to the intersection and a speed of $10$ m/s. A random number $N$ of other vehicles are positioned along the top and bottom roads with a random distance $p^{c,j}_o$ to the intersection, and a random desired speed $v_\mathrm{d}^j$. The other vehicles follow the Intelligent Driver Model (IDM)~\cite{idm2000}, with a set time gap of $t_\mathrm{d}^j=1$ s. One quarter of the vehicles stop at the intersection and three quarters continue through the intersection, regardless of the behavior of the ego vehicle. 
When a vehicle has passed the intersection and reached the end of the road, it is moved back to the other side of the intersection, which creates a constant traffic flow.
The simulator is updated at $25$ Hz, and decisions are taken at $4$ Hz.
The goal of the ego vehicle is to reach a position that is located $10$ m to the right of the last crossing point.


\begin{table}[!bt]
	\renewcommand{\arraystretch}{1.2}
	\caption{Parameters for simulator}
	\label{tab:simulator_parameters}
	\centering
	\begin{tabular}{lr}
		\toprule
		Number of vehicles, $N$ & $\{1,2,3,4\}$ \\
		Starting position ego, $p^{\mathrm{c},j}_\mathrm{e}$ & $[50, 60]$ m\\
		Starting position target, $p^{\mathrm{c},j}_\mathrm{o}$ & $[10, 55]$ m\\
        Desired velocity, $v_\mathrm{d}^j$ & $[8,12]$ m/s\\
		\bottomrule
	\end{tabular}

\end{table}

\subsection{MDP formulation\footnote{The full state is not directly observable, since the intentions of the surrounding vehicles are not known to the agent. Therefore, the problem is a Partially Observable Markov Decision Process (POMDP)~\cite{Kaelbling1998}. However, by using a $k$-Markov approximation, where the state consists of the $k$ last observations, the POMDP can be approximated as an MDP~\cite{Mnih2015}. For the scenarios that were considered in this study, it proved sufficient to simply use the last observation.}}
The following Markov decision process is used to model the decision-making problem.

\subsubsection{State space, $\mathcal{S}$}
\begin{figure}[!t]
    \centering
        \includegraphics[width=0.65\columnwidth]{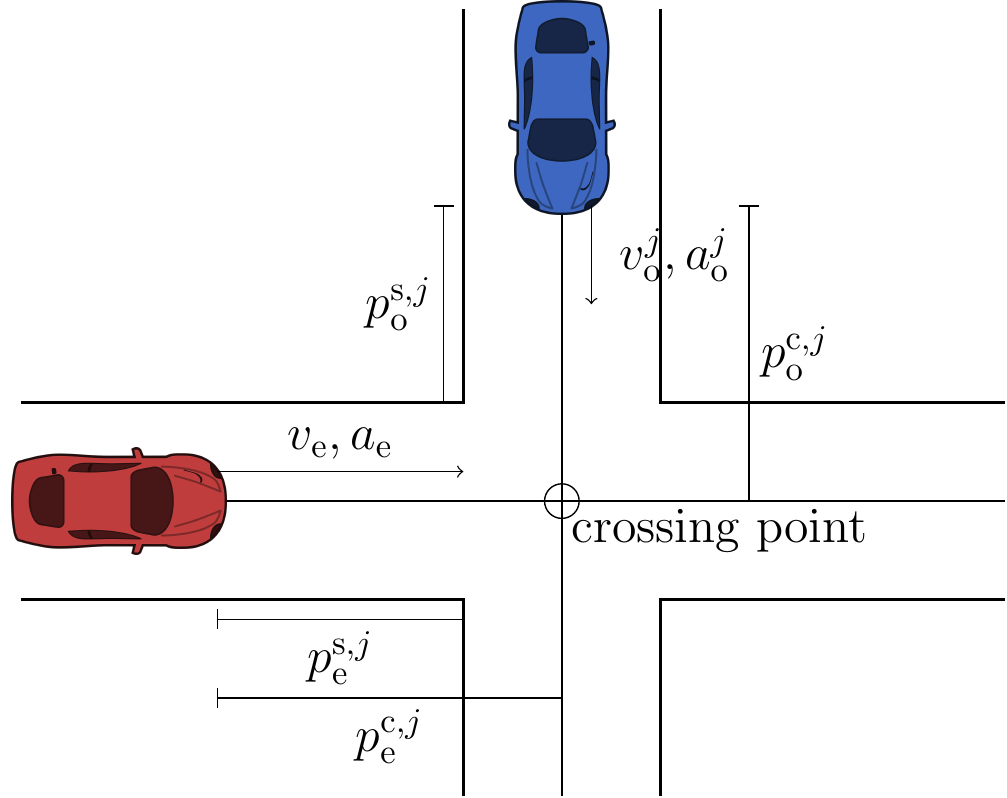}
        \caption{The state space definitions for a single crossing scenario, where subscript e and o denotes ego and other vehicle, respectively.}
    \label{fig:states}
\end{figure}

The design of the state of the system, 
\begin{align}
    s = (p^\mathrm{g}_\mathrm{e}, v_\mathrm{e}, a_\mathrm{e}, \{p^{\mathrm{s},j}_\mathrm{e}, p^{\mathrm{c},j}_\mathrm{e}, p^{\mathrm{s},j}_\mathrm{o}, p^{\mathrm{c},j}_\mathrm{o}, v^j_\mathrm{o}, a^j_\mathrm{o}\}_{j\in0,\dots,N}),
\end{align}
allows the description of intersections with different layouts~\cite{Tram2018}. The state, illustrated in Fig.~\ref{fig:states}, consists of the distance from the ego vehicle to the goal $p^\mathrm{g}_\mathrm{e}$, the velocity and acceleration of the ego vehicle, $v_\mathrm{e}$, $a_\mathrm{e}$, and the other vehicles, $v^j_\mathrm{o}$, $a_\mathrm{o}^j$, where $j$ denotes the index of the other vehicles. Furthermore, $p^{\mathrm{s},j}_\mathrm{e}$ and $p^{\mathrm{c},j}_\mathrm{e}$ are the distances from the ego vehicle to the start of the intersection and crossing point, relative to target vehicle $j$ respectively. The distances $p^{\mathrm{s},j}_\mathrm{o}$ and $p^{\mathrm{c},j}_\mathrm{o}$ are the distance from the other vehicles to the start of the intersection and the crossing point. 

\subsubsection{Action space, $\mathcal{A}$}
The action space consists of six tactical decisions: \{\textit{`take way', `give way', `follow car~$\{1, \dots , 4\}$'}\}, which set the target of the IDM controller. The \textit{`take way'} action treats the situation as an empty road, whereas the \textit{`give way'} action sets a target distance of $p^{s,j}_{e}$ and a target speed of $0$ m/s. The \textit{`follow car $j$'} actions sets the target distance to $p^{\mathrm{c},j}_\mathrm{e} - p^{\mathrm{c},j}_\mathrm{o}$ and target speed to $v^j_\mathrm{o}$. In cases where $p^{\mathrm{c},j}_\mathrm{o} > p^{\mathrm{c},j}_\mathrm{e}$, the target distance is set to a value that corresponds to timegap $0.5$ s.
The output of the IDM model is further limited by a maximum jerk $j_{\max}=5$ m/s\textsuperscript{3} and maximum acceleration $a_{\max}=5$ m/s\textsuperscript{2}. If less than four vehicles are present, the actions that correspond to choosing an absent vehicle are pruned by using Q-masking~\cite{Mukadam2017}.


\subsubsection{Reward model, $R$}
The objective of the agent is to reach the goal on the other side of the intersection, without colliding with other vehicles and for comfort reasons, with as little jerk $j_t$ as possible. Therefore, the reward at each time step $r_t$ is defined as
\begin{align*}
r_t = &\begin{cases}
1 & \text{at reaching the goal, }\\
-1 & \text{at a collision},\\
-\left(\frac{j_t}{j_{\max}}\right)^2\frac{\Delta \tau}{\tau_\mathrm{max}}         & \text{at non-terminating steps.}
\label{eq:reward}
\end{cases} 
\end{align*}
The non-terminating reward is scaled with the maximum time of an episode, $\tau_\mathrm{max}$, and the step time $\Delta \tau=0.04$ s, to ensure $\sum_{t=0}^{t=\tau_\mathrm{max}} \in [-1,0]$. Further details about the reward function can be found in previous research~\cite{tram2019}.


\subsubsection{Transition model, $T$}
The state transition probabilities are not known to the agent. However, the true transition model is defined by the simulation model, described in Sect.~\ref{sec:simulationModel}.

\subsection{Fallback action}
As mentioned in Sect.~\ref{sec:safetyCriterion}, a fallback action $a_{\mathrm{safe}}$ is used when $c_\mathrm{v}>c_\mathrm{v}^\mathrm{safe}$ for all available actions. This fallback action is set to \textit{`give way'}, with the difference that no jerk limitation is applied and with a higher acceleration limit $a_{\max} = 10$~m/s\textsuperscript{2}.

\subsection{Network architecture}
\begin{figure}[!t]
    \centering
        \includegraphics[width=0.99\columnwidth]{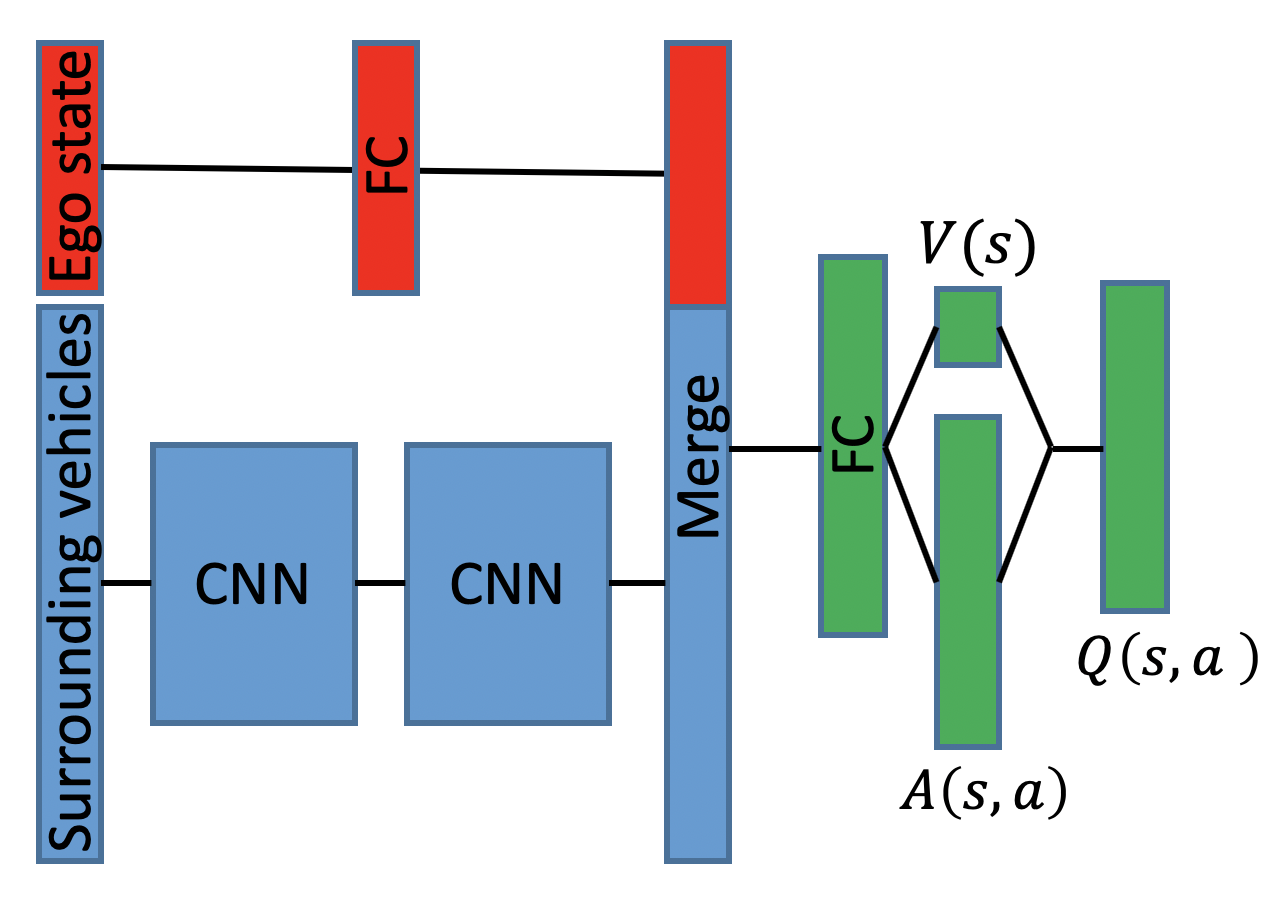}
        \caption{The neural network architecture that was used in this study.}
    \label{fig:neuralNetworkArchitecture}
\end{figure}

In previous studies, we have showed that a network architecture that applies the same weights to the input that describes the surrounding vehicles results in a better performance and speeds up the training process~\cite{Hoel2018},~\cite{Tram2018}. Such an architecture can be constructed by applying a one-dimensional convolutional neural network (CNN) structure to the surrounding vehicles' input. The network architecture that is used in this study is shown in Fig.~\ref{fig:neuralNetworkArchitecture}.
The first convolutional layer has $32$ filters, with size and stride set to six, which equals the number of state inputs of each surrounding vehicle, and the second convolutional layers has $16$ filter, with size and stride set to one. The fully connected (FC) layer that is connected to the ego vehicle input has $16$ units, and the joint fully connected layer has $64$ units. All layers use rectified linear units (ReLUs) as activation functions, except for the last layer, which has a linear activation function.
The final dueling structure of the network separates the estimation of the state value $V(s)$ and the action advantage $A(s,a)$~\cite{Wang2016}.
The input vector is normalized to the range $[\minus 1, 1]$. The input vector contains slots for four surrounding vehicles, and if less vehicles are present in the traffic scene, the empty input is set to $\minus 1$.

\subsection{Training process}

Algorithm~\ref{alg:training} is used to train the agent. The loss function of Double DQN is applied, which subtly modifies the maximization operation of Eq.~\ref{eq:loss} to $\gamma Q(s',\argmax_{a'} Q(s',a';\theta_i);\theta_i^-)$~\cite{Hasselt2016}. 
The Adam optimizer is used to update the weights~\cite{Kingma2014}, and $K$ parallel workers are used for the backpropagation step. The hyperparameters of the training process are shown in Table~\ref{tab:hyperparameters}, and the values were selected by an informal search, due to the computational complexity.

If the current policy of the agent decides to stop the ego vehicle, an episode could continue forever. Therefore, a timeout time is set to $\tau_\mathrm{max}=20$ s, at which the episode terminates. The last experience of such an episode is not added to the replay memory.
This trick prevents the agent to learn that an episode can end due to a timeout, and makes it seem like an episode can continue forever, which is important, since the terminating state due to the time limit is not part of the MDP~\cite{Hoel2018}.

\begin{table}[!bt]
	\renewcommand{\arraystretch}{1.2}
	\caption{Hyperparameters of Algorithm~\ref{alg:training} and baseline DQN.}
	\label{tab:hyperparameters}
	\centering
	\begin{tabular}{lr}
		\toprule
		Number of ensemble members, $K$ & $10$\\
		Prior scale factor, $\beta$ & $1$\\
		Experience adding probability, $p_\mathrm{add}$ & $0.5$\\
		Discount factor, $\gamma$ & $0.99$\\
		Learning start iteration, $N_\mathrm{start}$ & $50{,}000$\\
		Replay memory size, $M_\mathrm{replay}$ & $500{,}000$\\
		Learning rate, $\eta$ & $0.0005$\\
		Mini-batch size, $M_\mathrm{mini}$ & $32$\\
		Target network update frequency, $N_\mathrm{update}$ & $20{,}000$\\
		Huber loss threshold, $\delta$ & $10$\\

		\midrule

		Initial exploration constant, $\epsilon_\mathrm{start}$  & $1$\\
		Final exploration constant, $\epsilon_\mathrm{end}$ & $0.05$\\
		Final exploration iteration, $N_{\epsilon\mathrm{{\text -}end}}$ & $1{,}000{,}000$\\

		\bottomrule
	\end{tabular}
\end{table}

\subsection{Baseline method}
The Double DQN method, hereafter simply referred to as the DQN method, is used as a baseline. For a fair comparison, the same hyperparameters as for the ensemble RPF method is used, with the addition of an annealing $\epsilon$-greedy exploration schedule, which is shown in Table~\ref{tab:hyperparameters}. During test episodes, a greedy policy is used.

\section{Results}
\label{sec:results}

The results show that the ensemble RPF method outperforms the DQN method, both in terms of training speed and final performance, when the resulting agents are tested on scenarios that are similar to the training scenarios. When the fully trained ensemble RPF agent is exposed to situations that are outside of the training distribution, the agent indicates a high uncertainty and chooses safe actions, whereas the DQN agent collides with other vehicles. More details on the characteristics of the results are presented and briefly discussed in this section, whereas a more general discussion follows in Sect.~\ref{sec:discussion}. 

The ensemble RPF and DQN agents were trained in the simulated environment that was described in Sect.~\ref{sec:implementation}. After every $50{,}000$ training steps, the performance of the agents were evaluated on $100$ random test episodes. These test episodes were randomly generated in the same way as the training episodes, but kept fixed for all the evaluation phases. 

\subsection{Within training distribution}
\label{sec:resultsWithinDistribution}

\begin{figure}[!t]
    \centering
        \includegraphics[width=0.99\columnwidth]{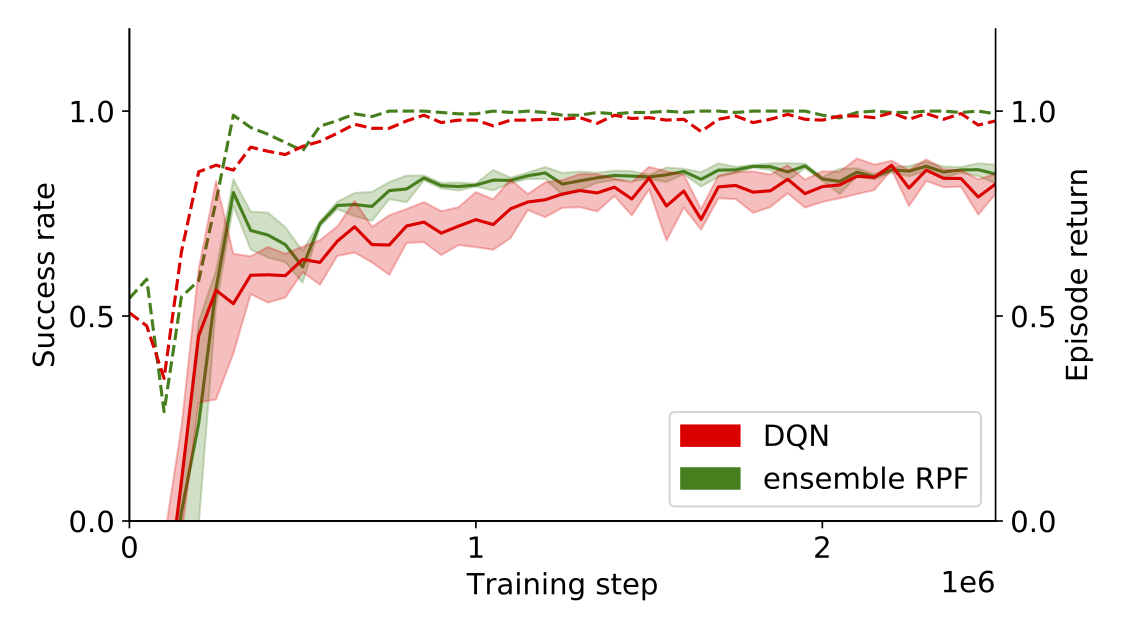}
    \caption{Proportion of test episodes where the ego vehicle reached its goal (dashed), and episode return (solid), over training steps for the ensemble RPF and DQN methods. The shaded areas show the standard deviation for $5$ random seeds.}
    \label{fig:returnAndSuccess}
\end{figure}

The average return and the average proportion of episodes where the ego vehicle reached the goal, as a function of number of training steps, is shown in Fig.~\ref{fig:returnAndSuccess}, for the test episodes. The figure also shows the standard deviation for $5$ random seeds, which generates different sets of initial parameters of the networks and different training episodes, whereas the test episodes are kept fixed. The results show that the ensemble RPF method both learns faster, yields a higher return, and causes less collisions than the DQN method. 
%

\begin{figure}[!t]
    \centering
        \includegraphics[width=0.99\columnwidth]{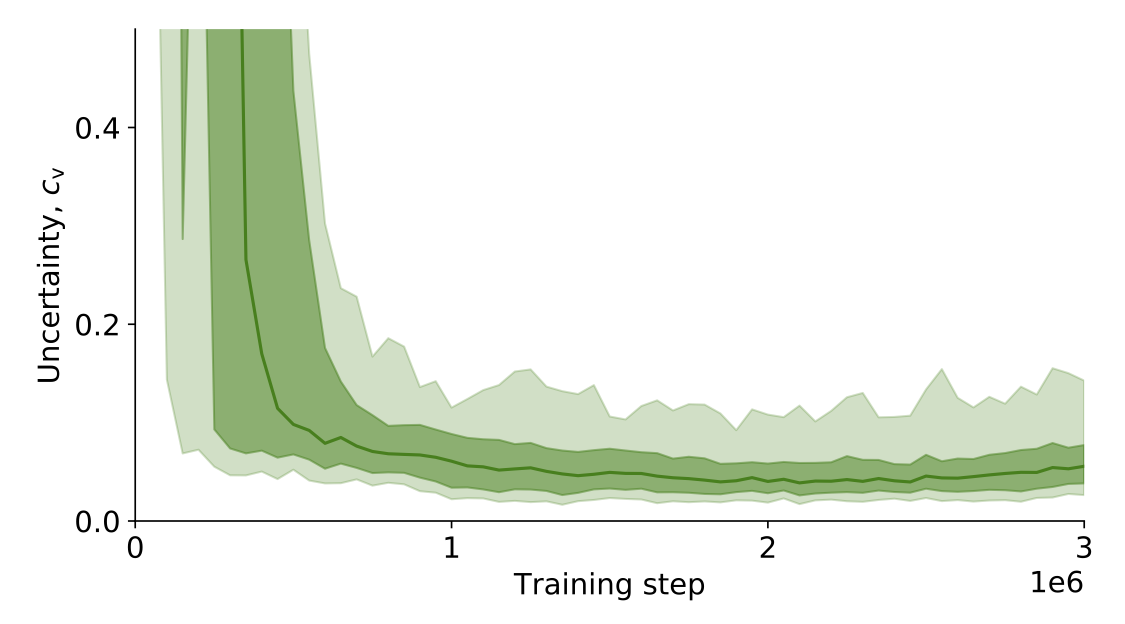}
        \caption{Mean coefficient of variation $c_\mathrm{v}$ for the chosen action during the test episodes. The dark shaded area shows percentiles $10$ to $90$, and the bright shaded area shows percentiles $1$ to $99$.}
    \label{fig:cv}
\end{figure}

Fig.~\ref{fig:cv} shows how the coefficient of variation $c_\mathrm{v}$ of the chosen action varies during the testing episodes. Note that the uncertainty of actions that are not chosen can be higher, which is often the case. After around one million training steps, the average value of $c_\mathrm{v}$ settles at around $0.04$, with a $99$ percentile value of $0.15$, which motivates the choice of setting $c_\mathrm{v}^\mathrm{safe}=0.2$.

\begin{figure}[!t]
    \centering
        \includegraphics[width=0.99\columnwidth]{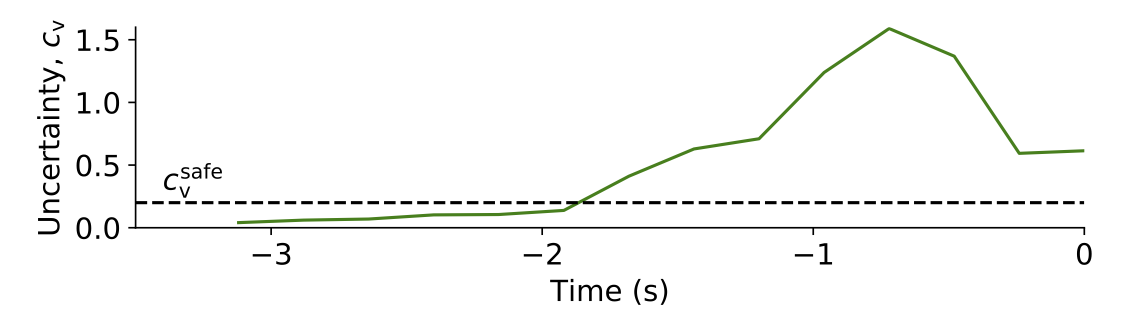}
        \caption{Uncertainty $c_\mathrm{v}$ during the time steps before one of the collisions in the test episodes, within the training distribution. The collision occurs at $t=0$ s.}
    \label{fig:cvDuringCrash}
\end{figure}

As shown in Fig.~\ref{fig:returnAndSuccess}, occasional collisions still occur during the test episodes when deploying the fully trained ensemble RPF agent. The reasons for these collisions are further discussed in Sect.~\ref{sec:discussion}. In one particular example of a collision, the agent fails to brake early enough and ends up in an impossible situation, where it collides with another vehicle in the intersection. However, the estimated uncertainty increases significantly during the time before the collision, when the incorrect actions are taken, see Fig.~\ref{fig:cvDuringCrash}. When applying the confidence criterion (Sect.~\ref{sec:safetyCriterion}), the agent instead brakes early enough, and can thereby avoid the collision. The confidence criterion was also applied to all the test episodes, which removed all collisions.

\subsection{Outside training distribution}

\begin{figure}[!t]
    \centering
    \begin{subfigure}[]{0.99\columnwidth}
    \centering
        \includegraphics[width=0.99\columnwidth]{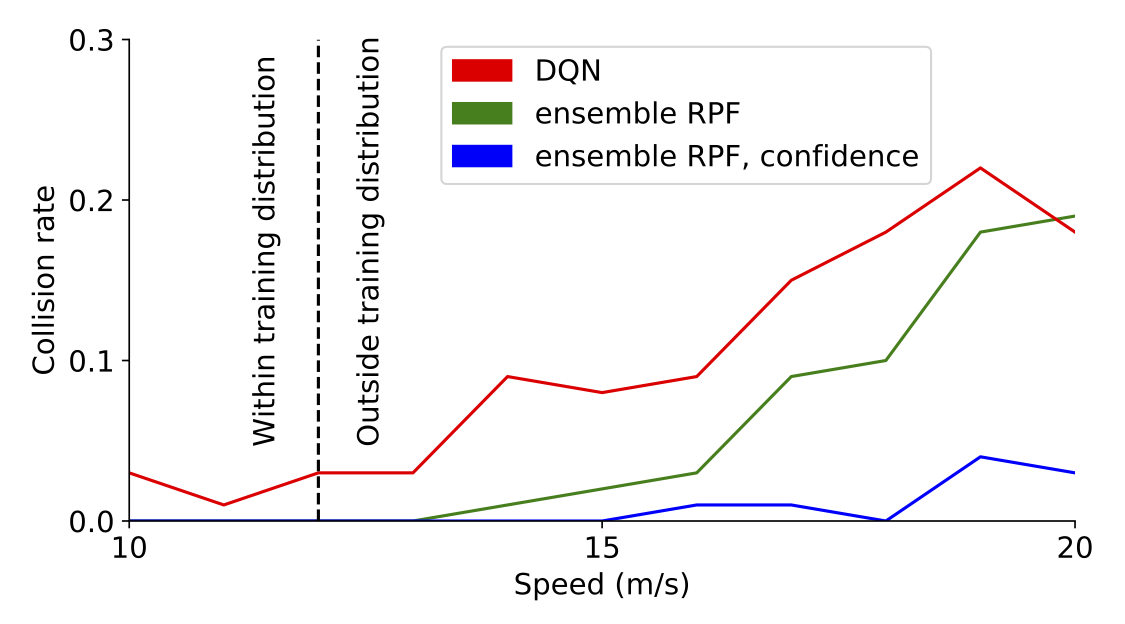}
        \caption{Proportion of collisions.}
    \end{subfigure}
    
    \vspace{5pt}
    
    \begin{subfigure}[]{0.99\columnwidth}
    \centering
        \includegraphics[width=0.99\columnwidth]{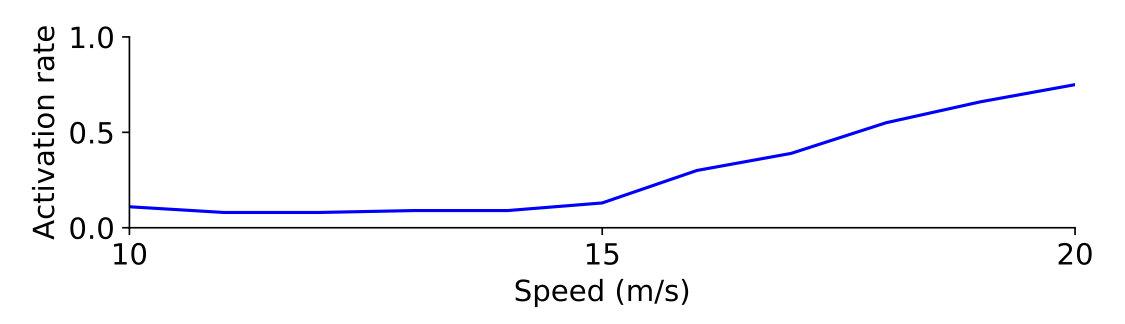}
        \caption{Proportion of episodes where $a_\mathrm{safe}$ was used at least once.}
    \end{subfigure}
    \caption{Performance of the ensemble RPF agent, with and without the confidence criterion, and the DQN agent, in test episodes with different set speeds $v_\mathrm{d}^j$ for the surrounding vehicles. 
    }
    \label{fig:performanceOutsideDistribution}
\end{figure}

The ensemble RPF agent that was obtained after three million training steps was tested in scenarios outside of the training distribution, in order to evaluate the agent's ability to detect unseen situations. The same testing scenarios as for within the distribution was used, with the exception that the speed of the surrounding vehicles was set to a single deterministic value, which was varied during different runs in the range $v_\mathrm{d}^j=[10,20]$ m/s. The proportion of collisions as a function of set speed of the surrounding vehicles is shown in Fig.~\ref{fig:performanceOutsideDistribution}, together with the proportion of episodes where the confidence criterion was violated at least once. The figure shows that when the confidence criterion is used, most of the collisions can be avoided. Furthermore, the violations of the criterion increase when the speed of the surrounding vehicles increase, i.e., the scenarios move further from the training distribution.

\begin{figure}[!t]
    \centering
    \begin{subfigure}[t]{0.48\columnwidth}
        \centering
        \includegraphics[width=0.7\columnwidth]{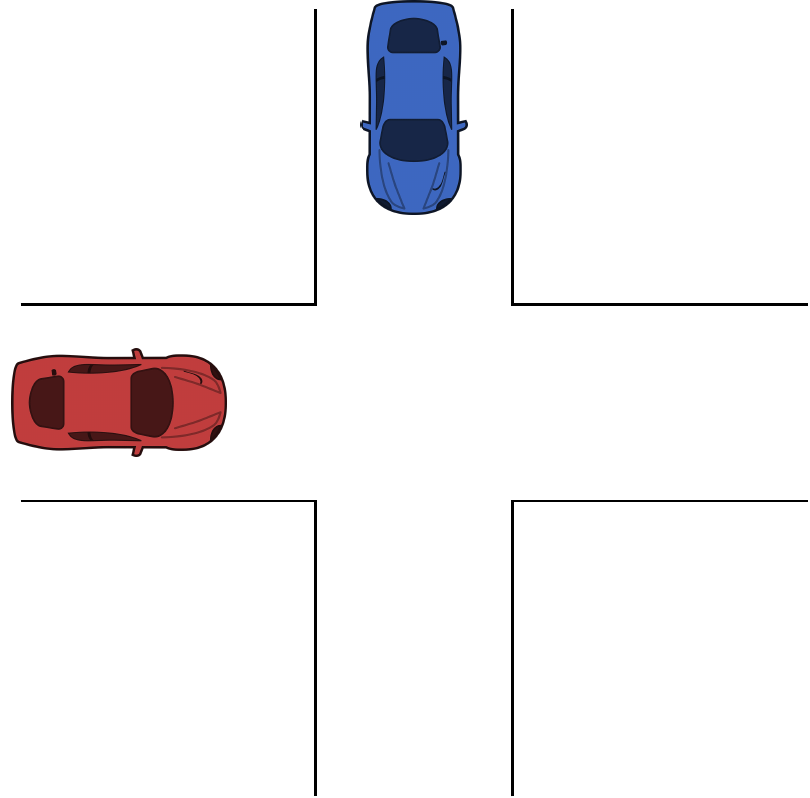}
        \caption{$t=0$, initial situation.}
    \end{subfigure}%
    ~ 
    \begin{subfigure}[t]{0.48\columnwidth}
        \centering
        \includegraphics[width=0.7\columnwidth]{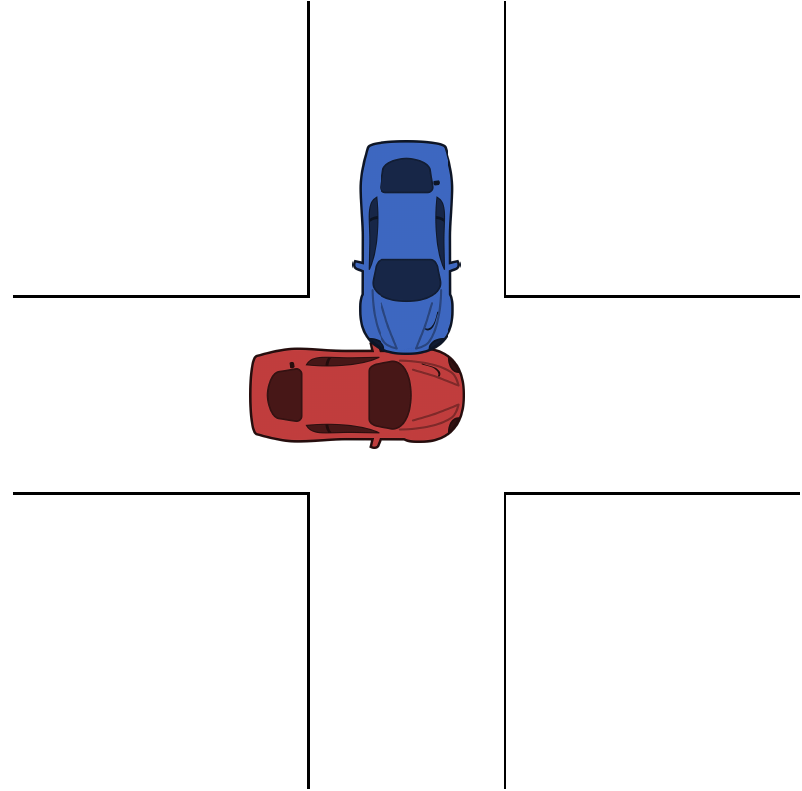}
        \caption{$t=1$, DQN and ensemble RPF without confidence criterion.}
    \end{subfigure}
    
    \vspace{5pt}
    
    \begin{subfigure}[t]{0.48\columnwidth}
        \centering
        \includegraphics[width=0.7\columnwidth]{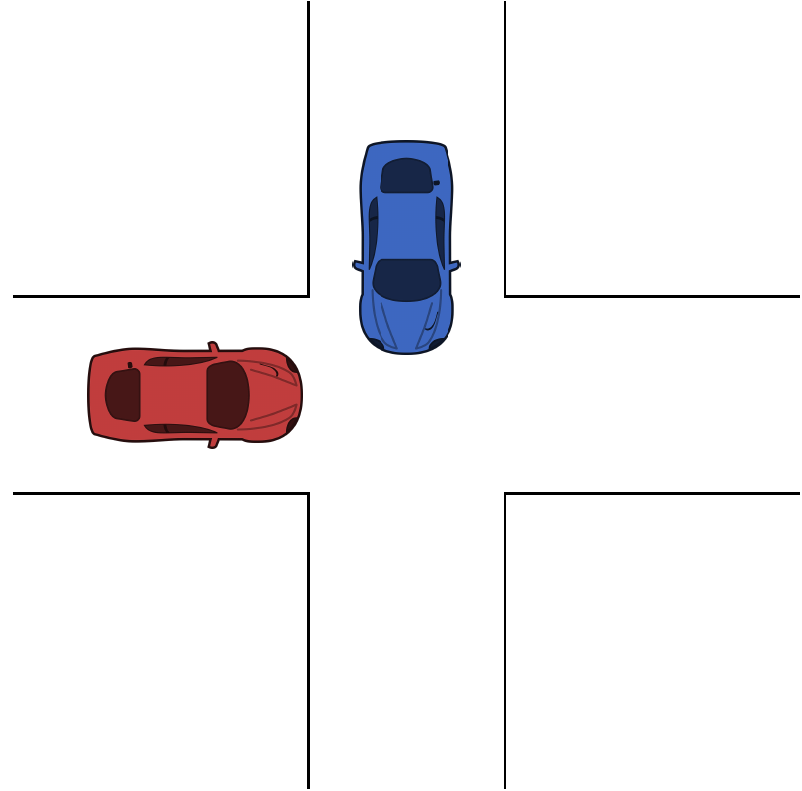}
        \caption{$t=1$, ensemble RPF with confidence criterion.}
    \end{subfigure}
    ~ 
    \begin{subfigure}[t]{0.48\columnwidth}
        \centering
        \includegraphics[width=0.7\columnwidth]{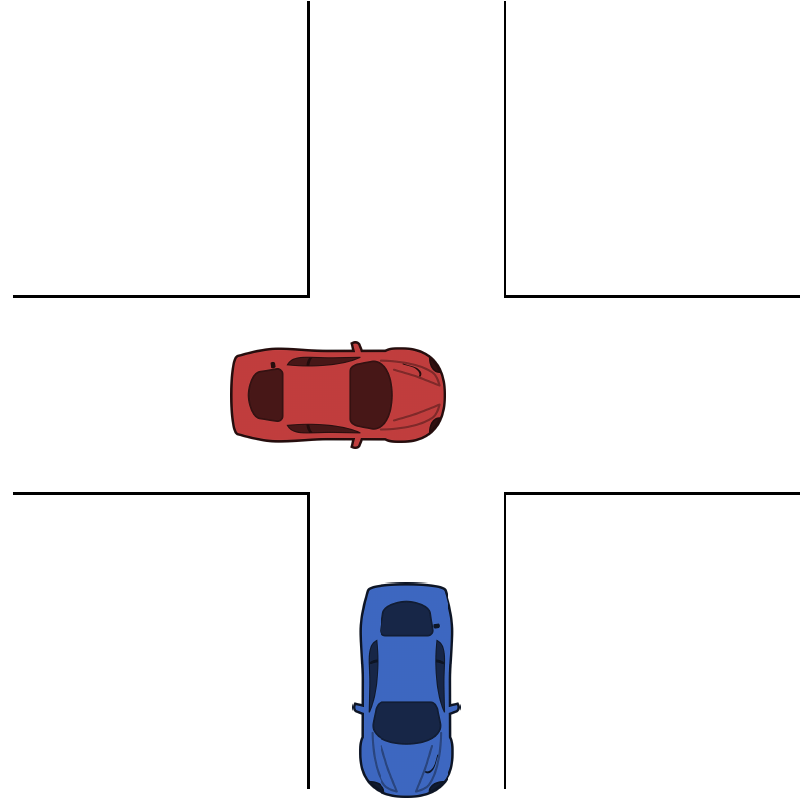}
        \caption{$t=1.5$, ensemble RPF with confidence criterion.}
    \end{subfigure}
    \caption{Example of a situation outside of the training distribution,
    where there would be a collision if the confidence criterion is not used. The vehicle at the top is here approaching the crossing at $20$ m/s.}
    \label{fig:collisionOutsideDistribution}

\end{figure}

An example of a situation that causes a collision is shown in Fig.~\ref{fig:collisionOutsideDistribution}, where an approaching vehicle drives with a speed of $20$ m/s. The $Q$-values of both the trained ensemble RPF and DQN agents indicate that the agents expect to make it over the crossing before the other vehicle. However, since the approaching vehicle drives faster than what the agents have seen during the training, a collision occurs. When the confidence criterion is applied, the uncertainty rises to $c_\mathrm{v}>c_\mathrm{v}^\mathrm{safe}$ for all actions when the ego vehicle approaches the critical region, where it has to brake in order to be able to stop, and a collision is avoided by choosing action $a_\mathrm{safe}$.

\section{Discussion}
\label{sec:discussion}






The results show that the ensemble RPF method can indicate an elevated uncertainty for situations that the agent has been insufficiently trained for, both within and outside of the training distribution.
In a previous study by the authors of this paper, we observed similar results when using the ensemble RPF method to estimate uncertainty outside of the training distribution in a highway driving scenario~\cite{Hoel2020}. In contrast, this paper shows that, in some cases, the ensemble RPF method can even detect situations with high uncertainty within the training distribution. Such situations include rare events that seldom or never occur during the training process, which makes it hard for the agent to provide an accurate estimate of the $Q$-values for the corresponding states. Since these states are seldom used to update the neural networks of the ensemble, the weights of the trainable networks will not adapt to the respective prior networks, and the uncertainty measure $c_\mathrm{v}$ will remain high for these rare events. This information is useful to detect edge cases within the training set and indicate when the decision of the trained agent is not fully reliable.

In this study, the estimated uncertainty is used to choose a safe fallback action if the uncertainty exceeds a threshold value. For the cases that are considered here, this confidence criterion removes all collisions within the training distribution, and almost all collisions when the speed of the surrounding vehicles is increased to levels outside of the training distribution. However, to guarantee safety by using a learning-based method is challenging, and an underlying safety layer is often used~\cite{Underwood2016}. The presented method could decrease the number of activations of such a safety layer, but possibly more importantly, the uncertainty measure could also be used to guide the training process to focus on situations that the current agent needs to explore further. Moreover, if an agent is trained in simulation and then deployed in real traffic, the uncertainty estimation of the agent could detect situations that should be added to the simulated world, in order to better match real-world driving.

The results show that the ensemble RPF method performs better and more stable than a standard DQN method within the training distribution. The main disadvantage is the increased computational complexity, since $K$ neural networks need to be trained. This disadvantage is somewhat mitigated in practice, since the design of the algorithm allows an efficient parallelization. Furthermore, the tuning complexity of the ensemble RPF and DQN methods are similar. Hyperparameters for the number of ensemble members $K$ and prior scale factor $\beta$ are introduced, but the parameters that control the exploration of DQN are removed.

\section{Conclusion}
\label{sec:conclusion}

The results of this study demonstrates the usefulness of using a Bayesian RL technique for tactical-decision making in an intersection scenario. The ensemble RPF method can be used to estimate the confidence of the recommended actions, and the results show that the trained agent indicates high uncertainty for situations that are outside of the training distribution. Importantly, the method also indicates high uncertainty for rare events within the training distribution. In this study, the confidence information was used to choose a safe action in situations with high uncertainty, which removed all collisions from within the training distribution, and most of the collisions in situations outside of the training distribution. 

The uncertainty information could also be used to identify situations that are not known to the agent, and guide the training process accordingly. To investigate this further is a topic for future work. Another subject for future work involves how to set the parameter value $c_v^\mathrm{safe}$ in a more systematic way, and how to automatically update the value during training.

\balance

\bibliographystyle{IEEEtran}
\bibliography{IEEEabrv,references}

\begin{thebibliography}{10}
\providecommand{\url}[1]{#1}
\csname url@samestyle\endcsname
\providecommand{\newblock}{\relax}
\providecommand{\bibinfo}[2]{#2}
\providecommand{\BIBentrySTDinterwordspacing}{\spaceskip=0pt\relax}
\providecommand{\BIBentryALTinterwordstretchfactor}{4}
\providecommand{\BIBentryALTinterwordspacing}{\spaceskip=\fontdimen2\font plus
\BIBentryALTinterwordstretchfactor\fontdimen3\font minus
  \fontdimen4\font\relax}
\providecommand{\BIBforeignlanguage}[2]{{%
\expandafter\ifx\csname l@#1\endcsname\relax
\typeout{** WARNING: IEEEtran.bst: No hyphenation pattern has been}%
\typeout{** loaded for the language `#1'. Using the pattern for}%
\typeout{** the default language instead.}%
\else
\language=\csname l@#1\endcsname
\fi
#2}}
\providecommand{\BIBdecl}{\relax}
\BIBdecl

\bibitem{NHTSA}
``Traffic safety facts,'' National Highway Traffic Safety Administration, Tech.
  Rep. DOT HS 812 261, 2014.

\bibitem{Sutton2018}
R.~S. Sutton and A.~G. Barto, \emph{Reinforcement Learning: An Introduction},
  2nd~ed.\hskip 1em plus 0.5em minus 0.4em\relax MIT Press, 2018.

\bibitem{Isele2018}
D.~{Isele}, R.~{Rahimi}, A.~{Cosgun}, K.~{Subramanian}, and K.~{Fujimura},
  ``Navigating occluded intersections with autonomous vehicles using deep
  reinforcement learning,'' in \emph{IEEE Int. Conf. on Robot. and Automat.
  (ICRA)}, 2018, pp. 2034--2039.

\bibitem{Tram2018}
T.~{Tram}, A.~{Jansson}, R.~{Gr\"{o}nberg}, M.~{Ali}, and J.~{Sj\"{o}berg},
  ``Learning negotiating behavior between cars in intersections using deep
  {Q}-learning,'' in \emph{IEEE Int. Conf. on Intell. Transp. Syst. (ITSC)},
  2018, pp. 3169--3174.

\bibitem{Wang2018}
P.~{Wang}, C.~{Chan}, and A.~d.~L.~{Fortelle}, ``A reinforcement learning based
  approach for automated lane change maneuvers,'' in \emph{IEEE Int. Veh. Symp.
  (IV)}, 2018, pp. 1379--1384.

\bibitem{Hoel2018}
C.~J. Hoel, K.~Wolff, and L.~Laine, ``Automated speed and lane change decision
  making using deep reinforcement learning,'' in \emph{IEEE Int. Conf. on
  Intell. Transp. Syst. (ITSC)}, 2018, pp. 2148--2155.

\bibitem{Shalev2016}
S.~Shalev{-}Shwartz, S.~Shammah, and A.~Shashua, ``Safe, multi-agent,
  reinforcement learning for autonomous driving,'' \emph{CoRR}, vol.
  abs/1610.03295, 2016.

\bibitem{Pan2017}
X.~Pan, Y.~You, Z.~Wang, and C.~Lu, ``Virtual to real reinforcement learning
  for autonomous driving,'' in \emph{Proc. of the Brit. Machine Vision Conf.
  (BMVC)}, 2017.

\bibitem{Bansal2018}
M.~Bansal, A.~Krizhevsky, and A.~S. Ogale, ``{ChauffeurNet}: Learning to drive
  by imitating the best and synthesizing the worst,'' \emph{Robot: Sci. \&
  Syst. (RSS)}, 2019.

\bibitem{Kendall2017}
A.~Kendall, V.~Badrinarayanan, and R.~Cipolla, ``Bayesian {SegNet}: Model
  uncertainty in deep convolutional encoder-decoder architectures for scene
  understanding,'' in \emph{Proc. of the Brit. Machine Vision Conf. (BMVC)},
  2017, pp. 57.1--57.12.

\bibitem{McAllister2017}
R.~McAllister \emph{et~al.}, ``Concrete problems for autonomous vehicle safety:
  Advantages of {Bayesian} deep learning,'' in \emph{Proc. of the 26th Int.
  Joint Conf. on Artif. Intell.}, 2017, pp. 4745--4753.

\bibitem{Kochenderfer2015}
M.~J. Kochenderfer, \emph{Decision Making Under Uncertainty: Theory and
  Application}.\hskip 1em plus 0.5em minus 0.4em\relax MIT Press, 2015.

\bibitem{Michelmore2018}
R.~Michelmore, M.~Kwiatkowska, and Y.~Gal, ``Evaluating uncertainty
  quantification in end-to-end autonomous driving control,'' \emph{CoRR}, vol.
  abs/1811.06817, 2018.

\bibitem{Dearden1998}
R.~Dearden, N.~Friedman, and S.~Russell, ``Bayesian {Q}-learning,'' in
  \emph{Proc. of the 15th Nat/10th Conf. on Artif. Intell./Innov. Appl. of
  Artif. Intell.}, 1998, p. 761–768.

\bibitem{Osband2018}
I.~Osband, J.~Aslanides, and A.~Cassirer, ``Randomized prior functions for deep
  reinforcement learning,'' in \emph{Adv. in Neural Inf. Process. Syst. 31},
  2018, pp. 8617--8629.

\bibitem{Hoel2020}
C.~J. Hoel, K.~Wolff, and L.~Laine, ``Tactical decision-making in autonomous
  driving by reinforcement learning with uncertainty estimation,'' submitted to
  IEEE Intell. Veh. Symp. (IV) 2020.

\bibitem{Watkins1992}
C.~J. C.~H. Watkins and P.~Dayan, ``Q-learning,'' \emph{Mach. Learn.}, vol.~8,
  no.~3, pp. 279--292, 1992.

\bibitem{Mnih2015}
V.~Mnih \emph{et~al.}, ``Human-level control through deep reinforcement
  learning,'' \emph{Nature}, vol. 518, no. 7540, pp. 529--533, 2015.

\bibitem{Garcia2015}
J.~Garc{{\'i}}a and F.~Fern{{\'a}}ndez, ``A comprehensive survey on safe
  reinforcement learning,'' \emph{J. of Mach. Learn. Res.}, vol.~16, no.~42,
  pp. 1437--1480, 2015.

\bibitem{Efron1982}
B.~Efron, \emph{The Jackknife, the Bootstrap and Other Resampling Plans}.\hskip
  1em plus 0.5em minus 0.4em\relax Soc. for Ind. and Appl. Math., 1982.

\bibitem{Osband2016}
I.~Osband, C.~Blundell, A.~Pritzel, and B.~Van~Roy, ``Deep exploration via
  bootstrapped {DQN},'' in \emph{Adv. in Neural Inf. Process. Syst. 29}, 2016,
  pp. 4026--4034.

\bibitem{tram2019}
T.~{Tram}, I.~{Batkovic}, M.~{Ali}, and J.~{Sjöberg}, ``Learning when to drive
  in intersections by combining reinforcement learning and model predictive
  control,'' in \emph{IEEE Int. Conf. on Intell. Transp. Syst. (ITSC)}, Oct
  2019, pp. 3263--3268.

\bibitem{idm2000}
M.~Treiber, A.~Hennecke, and D.~Helbing, ``Congested traffic states in
  empirical observations and microscopic simulations,'' \emph{Phys. Rev. E},
  vol.~62, pp. 1805--1824, 2000.

\bibitem{Kaelbling1998}
L.~P. Kaelbling, M.~L. Littman, and A.~R. Cassandra, ``Planning and acting in
  partially observable stochastic domains,'' \emph{Artif. Intell.}, vol. 101,
  no. 1-2, pp. 99--134, 1998.

\bibitem{Mukadam2017}
M.~Mukadam, A.~Cosgun, A.~Nakhaei, and K.~Fujimura, ``{Tactical Decision Making
  for Lane Changing with Deep Reinforcement Learning},'' \emph{Neural
  Information Processing Systems (NIPS)}, 2017.

\bibitem{Wang2016}
Z.~Wang, T.~Schaul, M.~Hessel, H.~Hasselt, M.~Lanctot, and N.~Freitas,
  ``Dueling network architectures for deep reinforcement learning,'' in
  \emph{Proc. of the 33rd Int. Conf. on Mach. Learn.}, vol.~48, 2016, pp.
  1995--2003.

\bibitem{Hasselt2016}
H.~van Hasselt, A.~Guez, and D.~Silver, ``Deep reinforcement learning with
  double {Q}-learning,'' in \emph{Proc. of the 39th AAAI Conf. on Artif.
  Intell.}, 2016, pp. 2094--2100.

\bibitem{Kingma2014}
D.~P. Kingma and J.~Ba, ``Adam: A method for stochastic optimization,''
  \emph{Int. Conf. on Learn. Repr.}, 12 2014.

\bibitem{Underwood2016}
S.~Underwood, D.~Bartz, A.~Kade, and M.~Crawford, ``Truck automation: Testing
  and trusting the virtual driver,'' in \emph{Road Veh. Automat. 3}, G.~Meyer
  and S.~Beiker, Eds.\hskip 1em plus 0.5em minus 0.4em\relax Springer, 2016,
  pp. 91--109.

\end{thebibliography}

\end{document}